\documentclass{article}

    \usepackage[nonatbib,final]{tackling_climate_workshop_style}

\usepackage[utf8]{inputenc} 
\usepackage[T1]{fontenc}    
\usepackage{url}            
\usepackage{booktabs}       
\usepackage{amsfonts}       
\usepackage{nicefrac}       
\usepackage{microtype}      
\usepackage{xcolor}
\usepackage{newclude}
\usepackage{glossaries}
\usepackage{graphicx}
\usepackage{wrapfig}
\usepackage{textcomp}
\usepackage[backend=biber, sorting=none]{biblatex}
\usepackage{caption, subcaption}
\usepackage{wrapfig}
\usepackage[labelfont=bf]{caption}
\usepackage{hyperref}       
\usepackage{amsmath}

\title{Heat Demand Forecasting with Multi-Resolutional Representation of Heterogeneous Temporal Ensemble}

\author{%
Satyaki Chatterjee$^{1,a,b^*}$, Adithya Ramachandran$^{1,a^*}$, Thorkil Flensmark B. Neergaard$^{2,c}$,\\
  \textbf{Andreas Maier}$^{3,a}$, \textbf{Siming Bayer}$^{4,a}$\\
  $^a$Pattern Recognition Lab, Friedrich-Alexander-Universität Erlangen-Nürnberg\\
  Martensstr. 3, 91058 Erlangen, Germany\\
  $^b$Diehl Metering GmbH, Donaustraße 120, 90451 Nürnberg, Germany\\
  $^c$Brønderslev Forsyning, Virksomhedsvej 20, 9700 Brønderslev, Denmark\\
  \texttt{$^1$satyaki.chatterjee@fau.de, $^1$adithya.ramachandran@fau.de,}\\
  {$^2$tbn@bronderslevforsyning.dk, $^3$andreas.maier@fau.de, $^4$siming.bayer@fau.de}\\
}

\newacronym{anns}{ANNs}{Artificial Neural Networks}
\newacronym{ann}{ANN}{Artificial Neural Network}
\newacronym{ai}{AI}{Artificial Intelligence}
\newacronym{rnn}{RNNs}{Recurrent Neural Networks}
\newacronym{sari}{SARIMAX}{Seasonal Auto-Regressive Integrated Moving Average with eXogenous factors}
\newacronym{cwt}{CWT}{Continuous Wavelet Transform}
\newacronym{cnn}{CNNs}{Convolutional Neural Networks}

\begin{document}

\maketitle
\def\thefootnote{*}\footnotetext{These authors contributed equally to this work}
\begin{abstract}
One of the primal challenges faced by utility companies is ensuring efficient supply with minimal greenhouse gas emissions. The advent of smart meters and smart grids provide an unprecedented advantage in realizing an optimised supply of thermal energies through proactive techniques such as load forecasting. In this paper, we propose a forecasting framework for heat demand based on neural networks where the time series are encoded as scalograms equipped with the capacity of embedding exogenous 
variables such as weather, and holiday/non-holiday.
Subsequently, CNNs are utilized to predict the heat load multi-step ahead. Finally, the proposed framework is compared with other state-of-the-art methods, such as SARIMAX and LSTM. The quantitative results from retrospective experiments show that the proposed framework consistently outperforms the state-of-the-art baseline method with real-world data acquired from Denmark. A minimal mean error of $7.54\%$ for MAPE and $417 kW$ for RMSE is achieved with the proposed framework in comparison to all other methods.

\end{abstract}
\section{Introduction}


\color{black}
In the current global scenario, the world is focused on gaining momentum in minimizing its carbon footprint. In $2020$, $90\%$ of the global heat supply was fueled by fossil fuels, and although significant technological advances have been made, the global energy requirement for space and water heating has been stable since $2010$ \cite{iea_heating}, \cite{iea_districtheating}. With an increasing urgency for conscious energy utilization, a reliable approach to predicting heat demand is imperative. The nature of the heat consumption data is in the form of a time series which is a set of data samples that provide information as a consequence of their sequential nature. Time series forecasting is the process of predicting target values at a future time period from observed historical data. The complexity of heat demand forecasting arises owing to its non-linear nature induced by human behavioral patterns, dependency on weather \cite{weatherUK}, working/non-working days \cite{calendar}, building properties \cite{buildings}, etc., imparting daily, weekly and seasonal patterns. Such complex dependencies make the heat demand forecasting problem multi-dimensional.


State-of-the-art heat forecasting methods can be categorized into three groups - statistical models, machine learning, and deep learning models. Statistical methods that are regression-based include Auto-regressive Moving Average (ARMA), Auto-regressive Integrated Moving Average (ARIMA), Seasonal ARIMA with eXogenous factors (SARIMAX), Auto-regressive Conditional Heteroskedasticity (ARCH) and their variants \cite{DOTZAUER2002277}, \cite{sarima}. Machine learning methods such as Support Vector Regression (SVR) are also used in the context of time series forecasting as standalone and or as hybrid models with statistical models \cite{svr}, \cite{ml}, \cite{chatterjee2021prediction}. Deep learning methods are currently at the forefront of time series forecasting as a consequence of their ability to learn non-linear functions through universal function approximators. Architectures such as \acrfull{rnn} and its progressive variants leverages the sequential nature of time series data to forecast future target values \cite{ann}, \cite{SURYANARAYANA2018141}, \cite{lstm}, \cite{lstm2}. 




An overlooked aspect of forecasting is the frequency component of the time series. Many real-life data are a complex aggregation of disparate components. 
With advances in signal processing, such time series signals can be transformed into a time-scale representation in the form of a scalogram using \acrfull{cwt}. A scalogram is analogous to an image that represents the coefficients of \acrshort{cwt} in time and scale as its two spatial coordinates with detailed time-frequency resolution compared to a spectrogram which is limited with fixed window size, thus enabling a multi-resolution feature analysis of the time series. The scales are inversely proportional to frequencies. This essentially shifts the domain from time series to computer vision where Convolutional Neural Networks (CNNs) are employed for image classification, and object recognition among others. We leverage this image-like representation of the time series to learn localised time-frequency features, discerning different frequencies at different points of time \cite{cwt}, \cite{wt}. 


Considering the multi-faceted forecasting challenge, we propose a framework contributing the following: (1) A deep learning based forecasting model for short-term 24 hours ahead heat demand at a district level; (2)A multi-resolution representation (Scalogram) based framework capturing localized non-linear time-frequency features of heat demand with exogenous variables; (3) Capacity of the framework to forecast in different seasons and evaluate its performance with existing standard deep learning and statistical method.



\section{Method Overview}

\begin{figure}[htbp]
\centering
\includegraphics[width=1.0\linewidth]{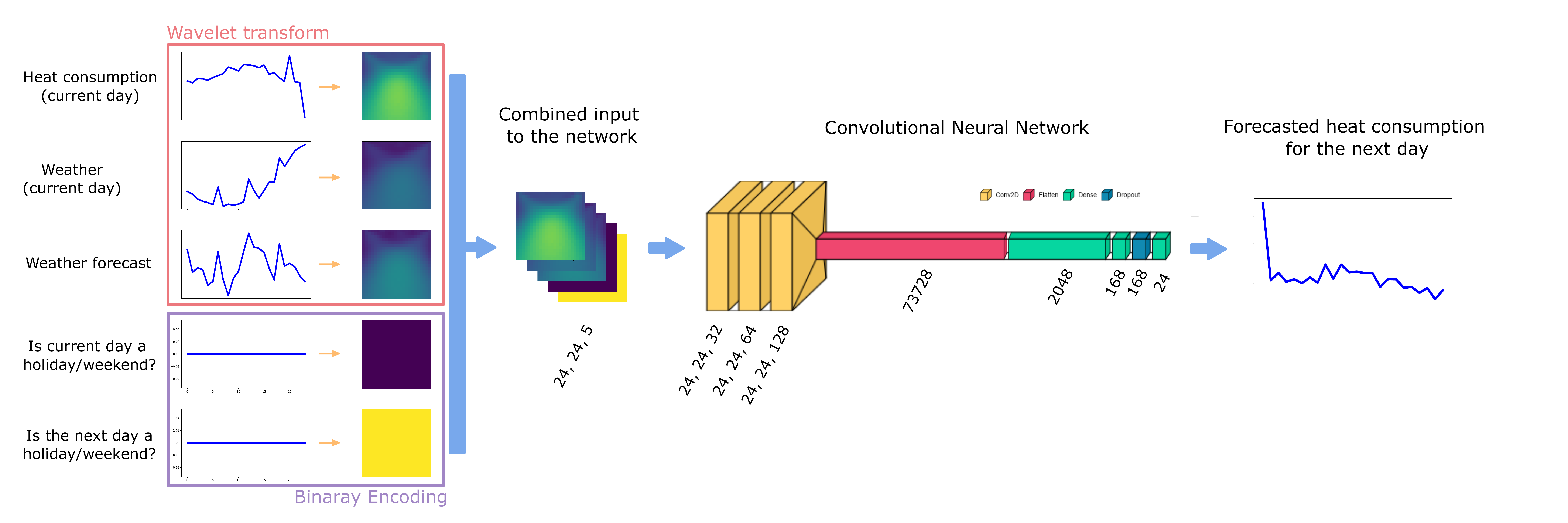}
\caption{Pictorial representation of the framework.}
\label{Fig:block_diagram}
\end{figure}

The core methodology of our framework (Figure \ref{Fig:block_diagram}) relies upon multi-step sequence-to-sequence prediction with the ability to process multi-resolution representation (Wavelet Scalograms) of multiple time series inputs, viz. historical consumption, historical and forecasted weather data, and encoded exogenous information viz. day of the week or a day being public holiday concurrently.

Let \(\mathbf{x}(t) = [x_{t-h}, x_{t-h+1}, ..., x_t]\) where \(x_i \in \mathbb{R}\) is the consumption value at time \(i\) represent the \(h\) historical observations that are leveraged to forecast the target variables \(\mathbf{y}(t) = [x_{t+1}, x_{t+2}, ..., x_{t+n}]\) where \(n\) is the forecasting horizon. The wavelet scalograms for historical heat consumption, historical weather \(\mathbf{w_{p}}(t) = [w_{t-h}, w_{t-h+1}, ..., w_t]\), and forecasted weather over the forecasting horizon \(\mathbf{w_{f}}(t) = [w_{t+1}, w_{t+2}, ..., w_{t+n}]\) are generated through \acrshort{cwt}. The wavelet transform converts each signal from a 1-dimensional time series into multi-dimensional data of size \(s \times h\), where $s$ is the number scales in the scalogram. The mathematical background for \acrshort{cwt} is illustrated in \cite{wavelet_mother}, \cite{wavelet_transform}.
The individual wavelet scalograms are of the same dimension due to a constraint on $s$ to be constant for all three data streams to enable concatenation for a 3-channel image-like representation of size \(3 \times s \times h\). The use of \(\mathbf{w_{p}}(t)\) and \(\mathbf{w_{f}}(t)\) places an additional constraint such that \(h=n\). Additionally, the weekday/holiday (or weekend) information of the current day and the next day are encoded into a matrix of dimension \(s \times h\) having only ones or only zeros, depending upon whether the day of concern is a weekend/public holiday or a weekday. These two matrices are further concatenated to the existing image-like 3-channel data, to form a 5-channel input to the \acrshort{cnn}. The CNN based model is trained with \((N/h)\) number of examples where the model aims to learn and translate \(h\) historical observations into \(n\) future estimations, by extracting features like the trend, multiple seasonality, dependencies on external factors and latent features from the multi-channel input. 
The pooling layer after convolution is deactivated because the pooling layers tend to smoothen the predictions \cite{pooling}. The CNN model which is used here has three convolution layers followed by one flatten layer and three fully-connected layers with a dropout between the last two fully-connected layers. The output layer has \(n\) number of nodes, the same as the forecasting horizon (\(n\) hours).

\section{Experimental setup}

Meter-level heat consumption data from $2015$ to $2018$ of a Danish utility with three district heating zones are utilized to conduct retrospective experiments. The consumption data of meters are hourly sampled and are aggregated at a zonal level to give district heating demand. Weather data is represented by feel-like temperature with an hourly resolution for the same Danish town as feel-like temperature conveys combined information of maximum and minimum temperature, wind speed, humidity, and other meteorological information. We also incorporated the information on public holiday from the Danish calendar along with the information about the day of the week. The consumption data, weather data, and the day of week/holiday data are sampled from the same time period. This data is then split into $730$ days for training, $180$ days for both validation and testing. The input to the model is a 5-channel image-like data where the scalograms are generated with \(h, s = 24\) using the Mexican-hat wavelet basis and the output is the forecasted heat demand with \(n=24\). 
In order to prevent the model from being trained with erroneous observations, the accumulated consumption data from the meters are first checked for monotonic increase as negative consumption or flow is not plausible. Anomalous data indicating a negative rate of consumption are adjusted to a zero consumption state. The actual heat consumption data is obtained through a first-order difference of the accumulated consumption data. The actual consumption data, along with the feel-like temperature are further feature scaled (normalised) between zero and one.

The three convolution layers have \([32, 64, 128]\) kernels with a kernel size of three, with 'same' padding and Rectified Linear Unit (ReLU) activation function whereas the fully connected layers are activated though Leaky-ReLU activation function. We use the Adam optimizer and Mean-Squared-Error (MSE) as a loss function for training the model. The network is trained till convergence and a suitable model is selected according to the validation loss to prevent over-fitting. A Batch size of 7 was chosen for training so that each batch represents each week. For quantitative evaluation, we use the Mean Absolute Percentage Error (MAPE) metric for $24$-hours ahead prediction. Due to the skewness and asymmetric tendency of penalization of MAPE in time series prediction \cite{Goodwin},\cite{chatterjee2021prediction}, Root Mean Squared Error (RMSE) is additionally chosen to evaluate the forecasting performance.



\section{Results and Discussion}

\begin{figure}[htbp] 
    \centering
    \begin{subfigure}[t]{0.48\linewidth} 
        \includegraphics[width=\linewidth]{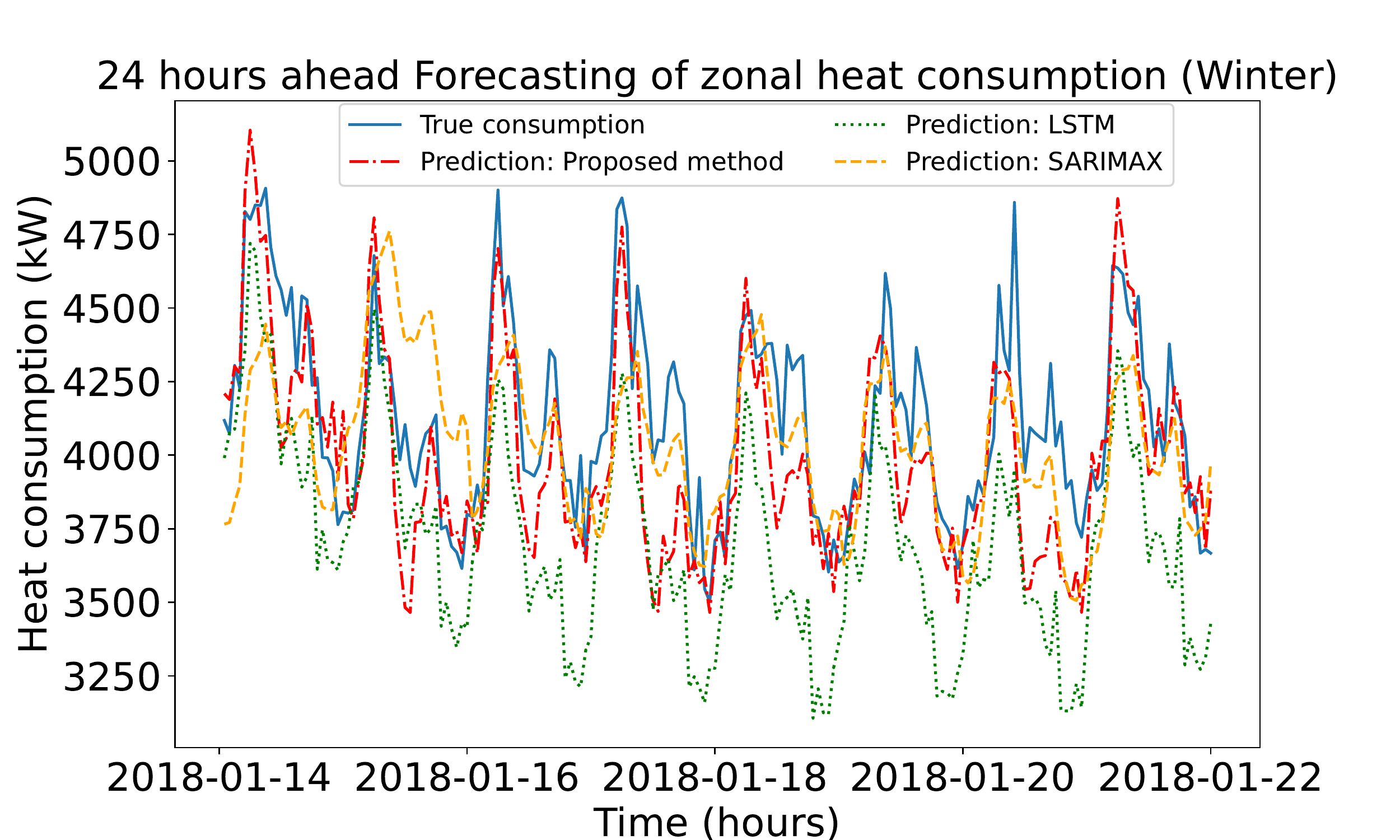} 
    \end{subfigure}
    \begin{subfigure}[t]{0.48\linewidth} 
        \includegraphics[width=\linewidth]{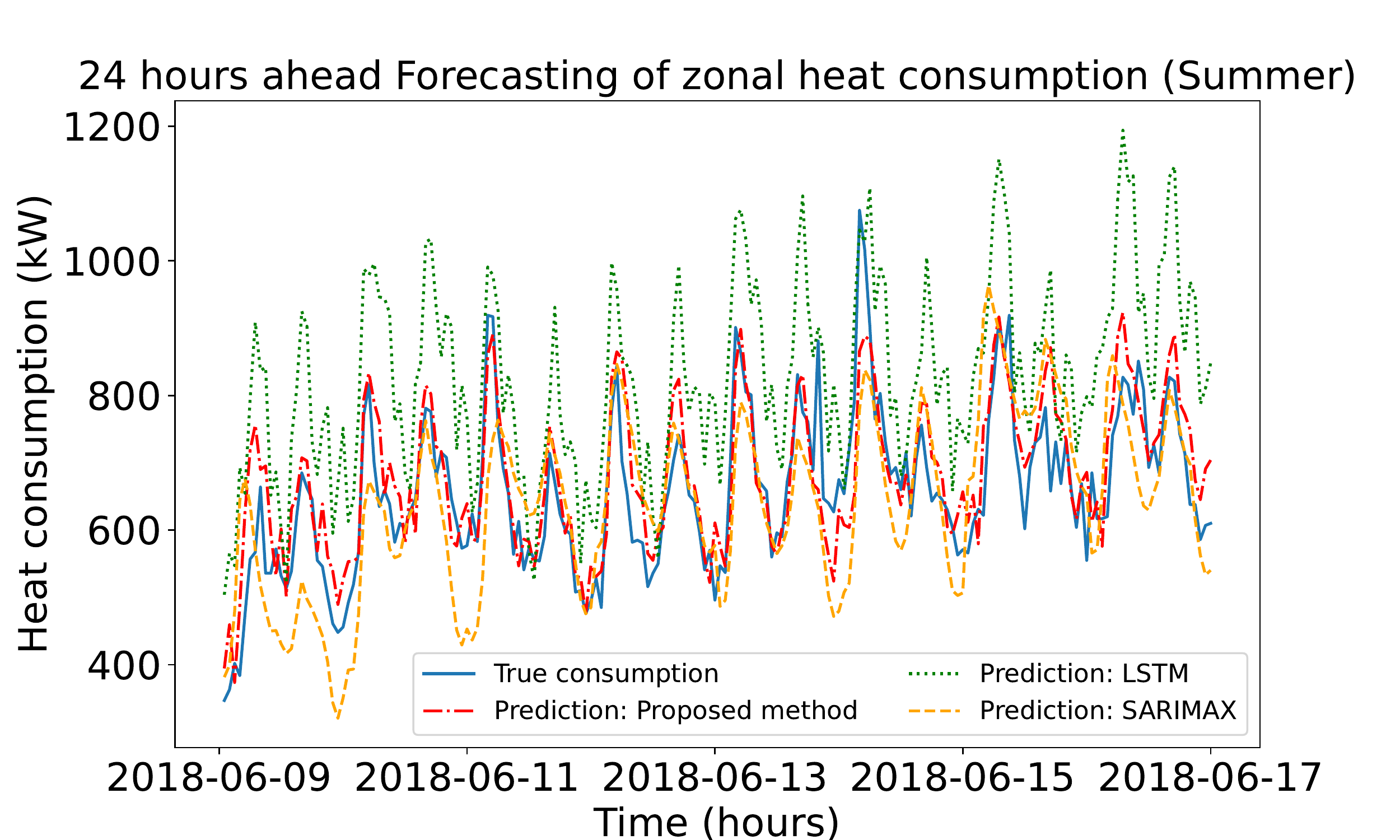} 
        \label{Fig:summerplot}
    \end{subfigure}
    \begin{subfigure}[t]{0.48\linewidth} 
        \includegraphics[width=\linewidth]{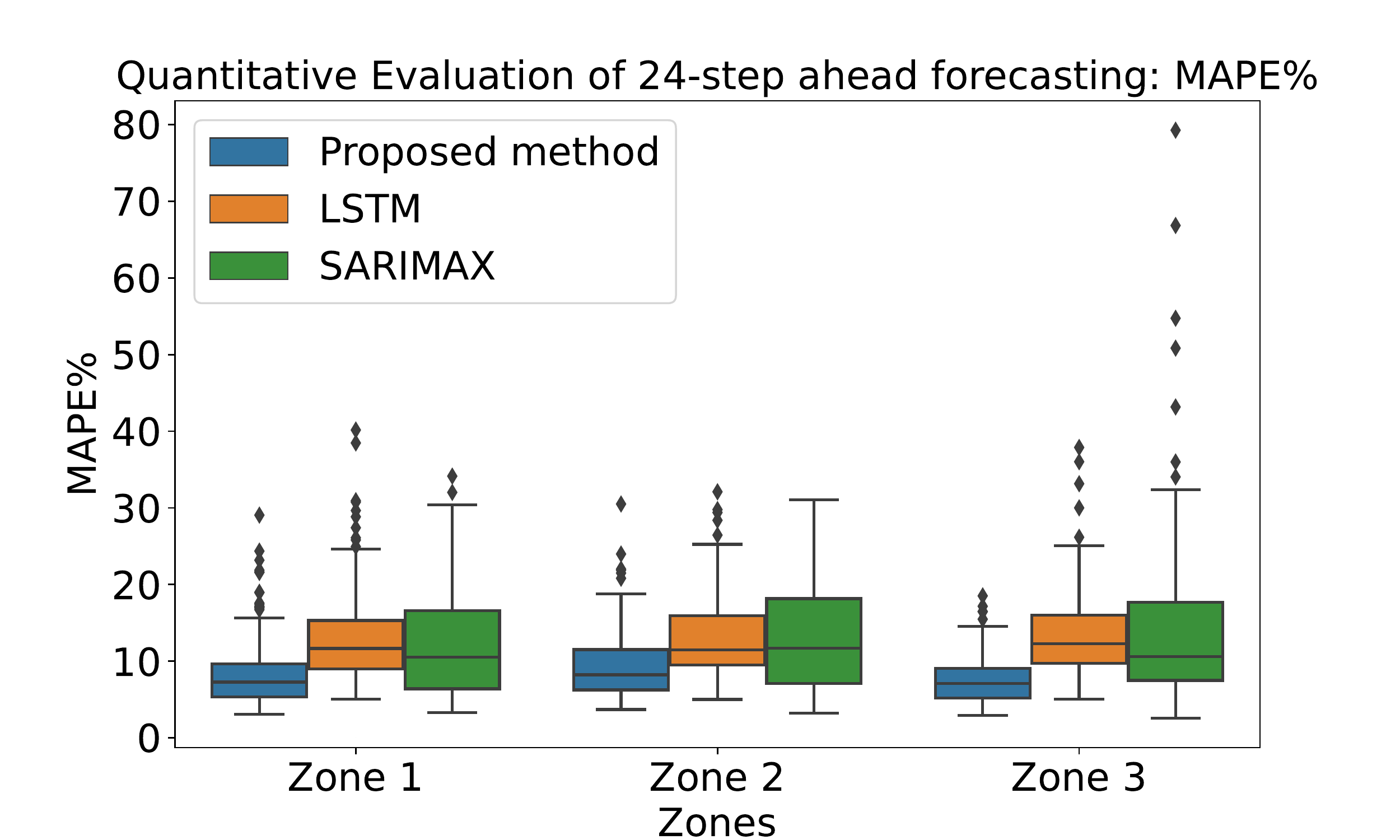} 
    \end{subfigure}
    \begin{subfigure}[t]{0.48\linewidth} 
        \includegraphics[width=\linewidth]{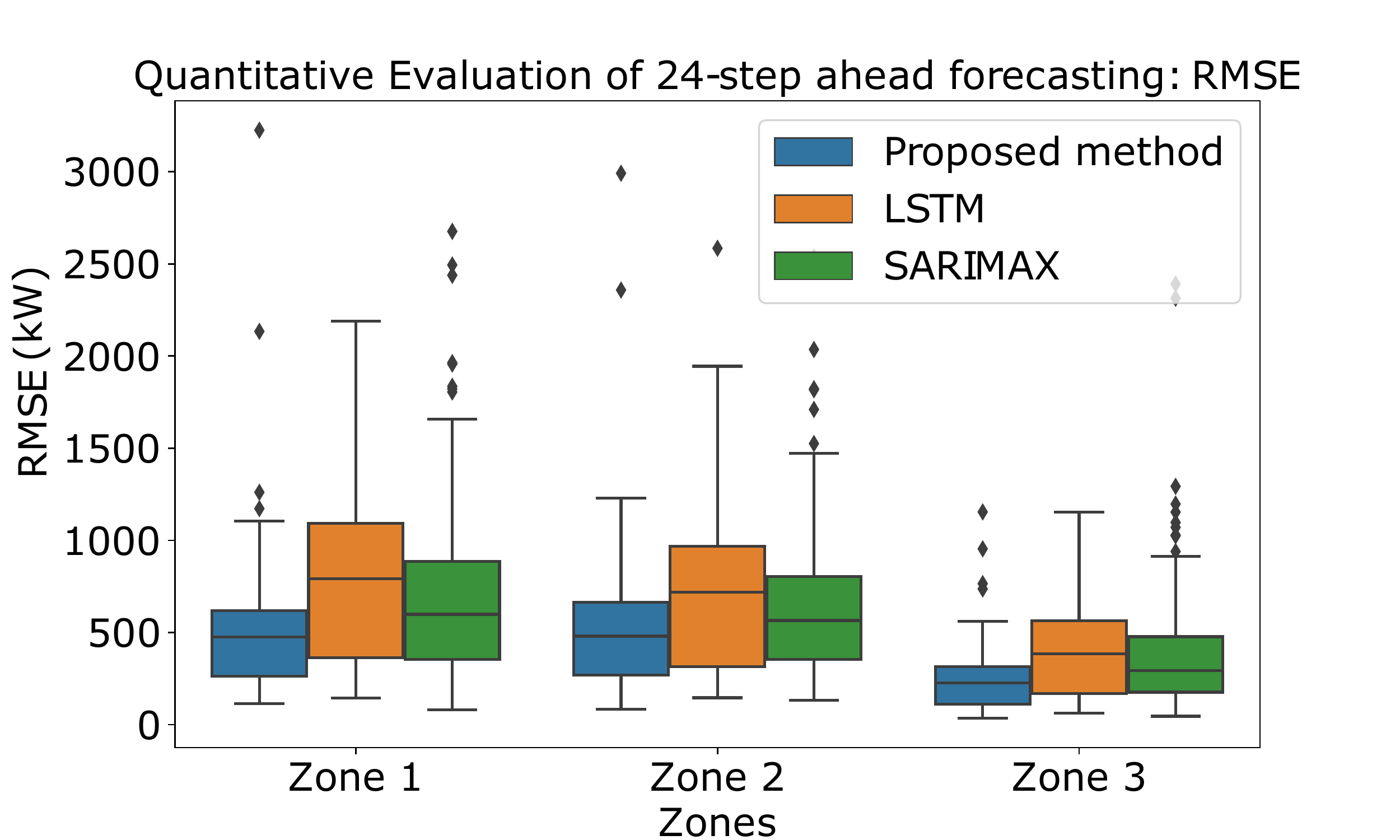} 
    \end{subfigure}
    \caption{Qualitative comparative evaluation of all models over eight days in winter $2018$ (top left) and summer $2018$ (top right) in a heat supply zone; MAPE of prediction over different zones (bottom left), RMSE of prediction over different zones (bottom right).}
    \label{Fig:metrics_plots}
\end{figure}

The performance of the proposed framework is evaluated against an LSTM model and \acrshort{sari}, with a constant experimental setup, across different supply zones, and over different climatic seasons, both qualitatively and quantitatively. Figure \ref{Fig:metrics_plots} illustrates the performances of the models across different seasons of the year in different zones. 
The qualitative comparison depicted in Figure \ref{Fig:metrics_plots} (top left, top right) 
demonstrates both the superiority of our proposed framework in capturing the time-varying trend present in daily consumption over the week or season as well as the inherent capacity to capture the underlying daily level load profile regardless of the season. In contrast, the LSTM model shows its ability to predict daily level load profiles across different seasons but fails to capture the change in the long-term context of a global trend across different seasons leading to under forecasting in winter and over forecasting in summer. The SARIMAX model learns the global season-wise trend and a simplified daily seasonality but not the extreme dynamic range of consumption during weekends or public holidays. The proposed method, in comparison, demonstrates its ability to not only learn the global trend and daily seasonality, but also its relationship with other factors like feel-like temperature, and holiday information, along with the local fluctuation of consumption that happens in a day.

  This proves the ability of our proposed model to learn the dependency on both frequency and temporal components, which is present in the wavelet scalograms. The quantitative evaluations shown in Figure \ref{Fig:metrics_plots} (bottom left and bottom right)  
  depicts the superiority of our proposed method over the other two baseline methods in terms of its lower mean MAPE, and much lower variance across different seasons and zones, which proves our claim regarding the method's robustness to comprehend consumption patterns irrespective of seasons or district heating zones.

\section{Conclusion}
In this work, we formulate a CNN-based framework for heat load prediction with an image-like representation for time series.
As a consequence, exogenous variables such as weather data, information regarding workday/weekends/holidays can be incorporated as additional channels in a multi-dimensional image straightforwardly.
Using the proposed framework, we perform a 24 hours ahead forecast for hourly sampled data and compare its performance to a baseline LSTM model as well as the SARIMAX method. Both qualitative and quantitative results demonstrate the ability of the proposed framework for heat demand forecast in comparison to state-of-the-art methods with lower forecasting error metrics. One limitation in the proposed framework is that the forecasting horizon and historical inputs must be of the same size to enable concatenation to produce the 5-channel input. As an extension to this work, the effects of convolution size, type of convolution can be explored to understand the importance of spatial features in the scalogram. From an application standpoint, under-forecasting heat demand is less preferred than over-forecasting, to ensure adequate supply. In this regard, a loss function that strongly penalises under forecasting than over forecasting needs to be incorporated. Forecasting at a district zonal level helps with optimizing the distribution of heat produced at a zonal level, enabling informed decision making - leading to further reduction of carbon.


\printbibliography 

\newpage
\section{Appendix}

\subsection{Wavelet Scalogram}
\begin{figure}[h] 
    \centering
    \begin{subfigure}[h]{0.45\linewidth} 
        \centering
        \includegraphics[width=\linewidth]{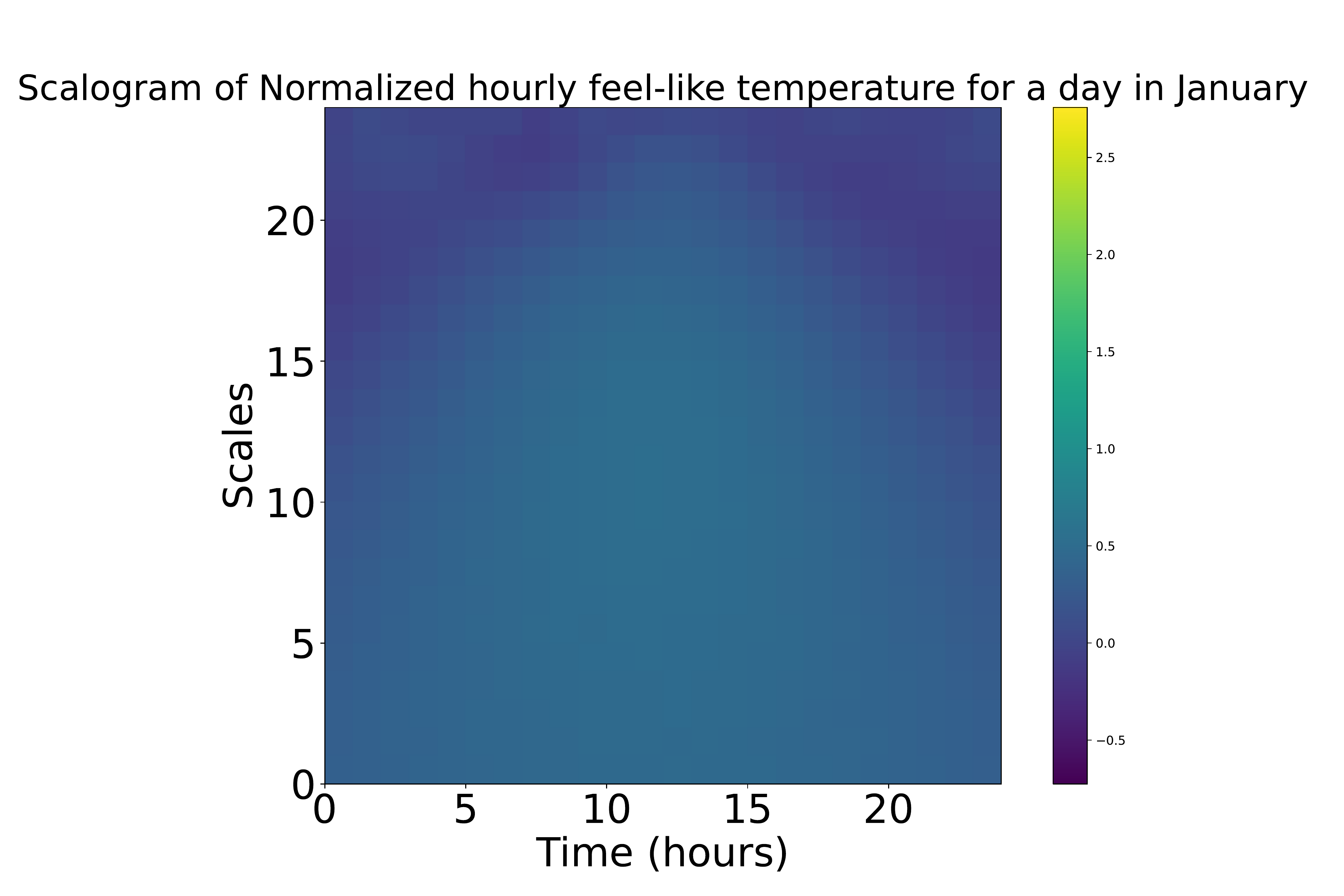} 
        \caption{Wavelet scalogram of feel-like temperature for a day in January.} 
        \label{Fig:weather_scalo_jan}
    \end{subfigure}
    \begin{subfigure}[h]{0.45\linewidth} 
        \centering
        \includegraphics[width=\linewidth]{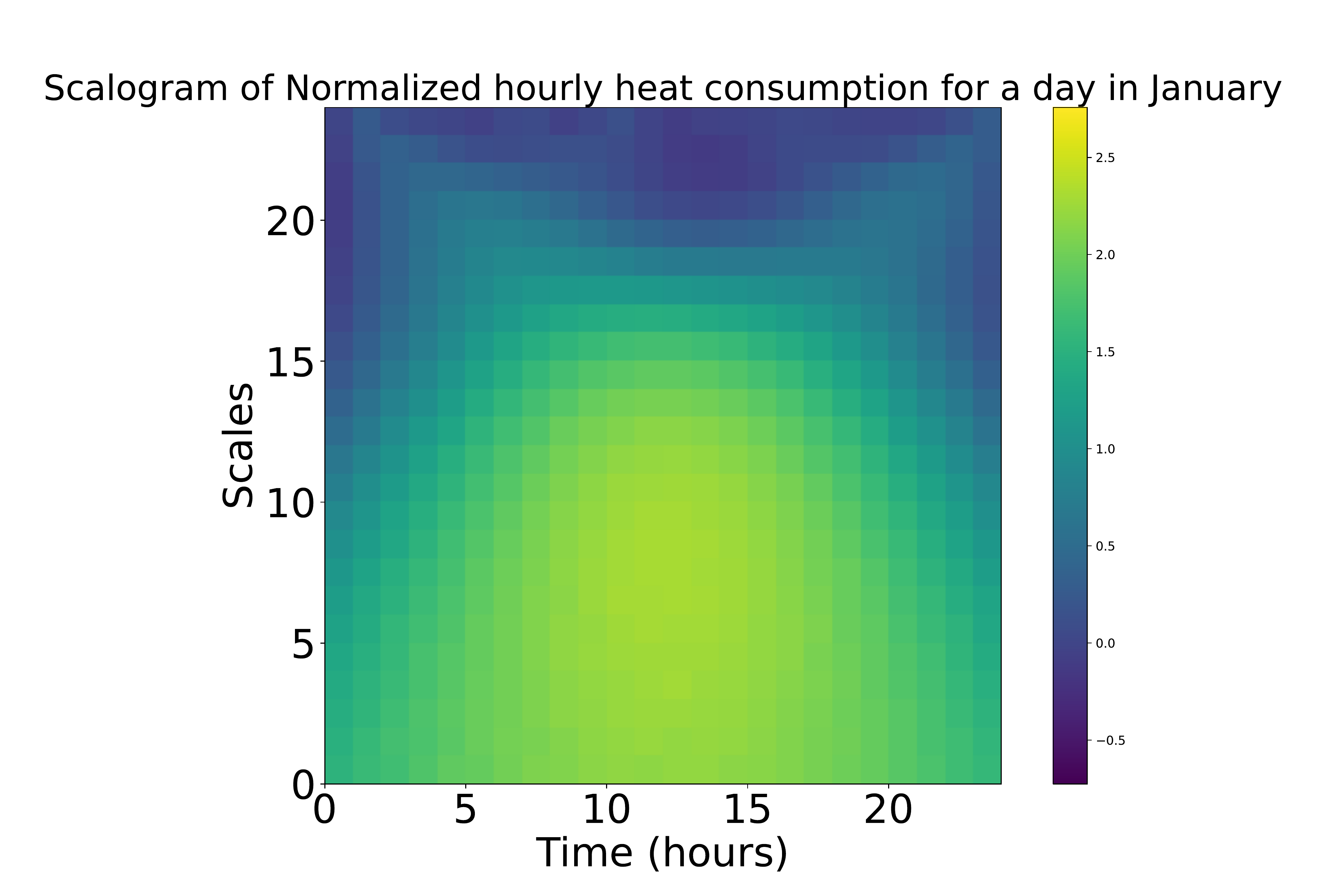} 
        \caption{Wavelet scalogram of heat consumption for a day in January.} 
        \label{Fig:Scalo_heat_jan}
    \end{subfigure}
    \centering
    \begin{subfigure}[h]{0.45\linewidth} 
        \centering
        \includegraphics[width=\linewidth]{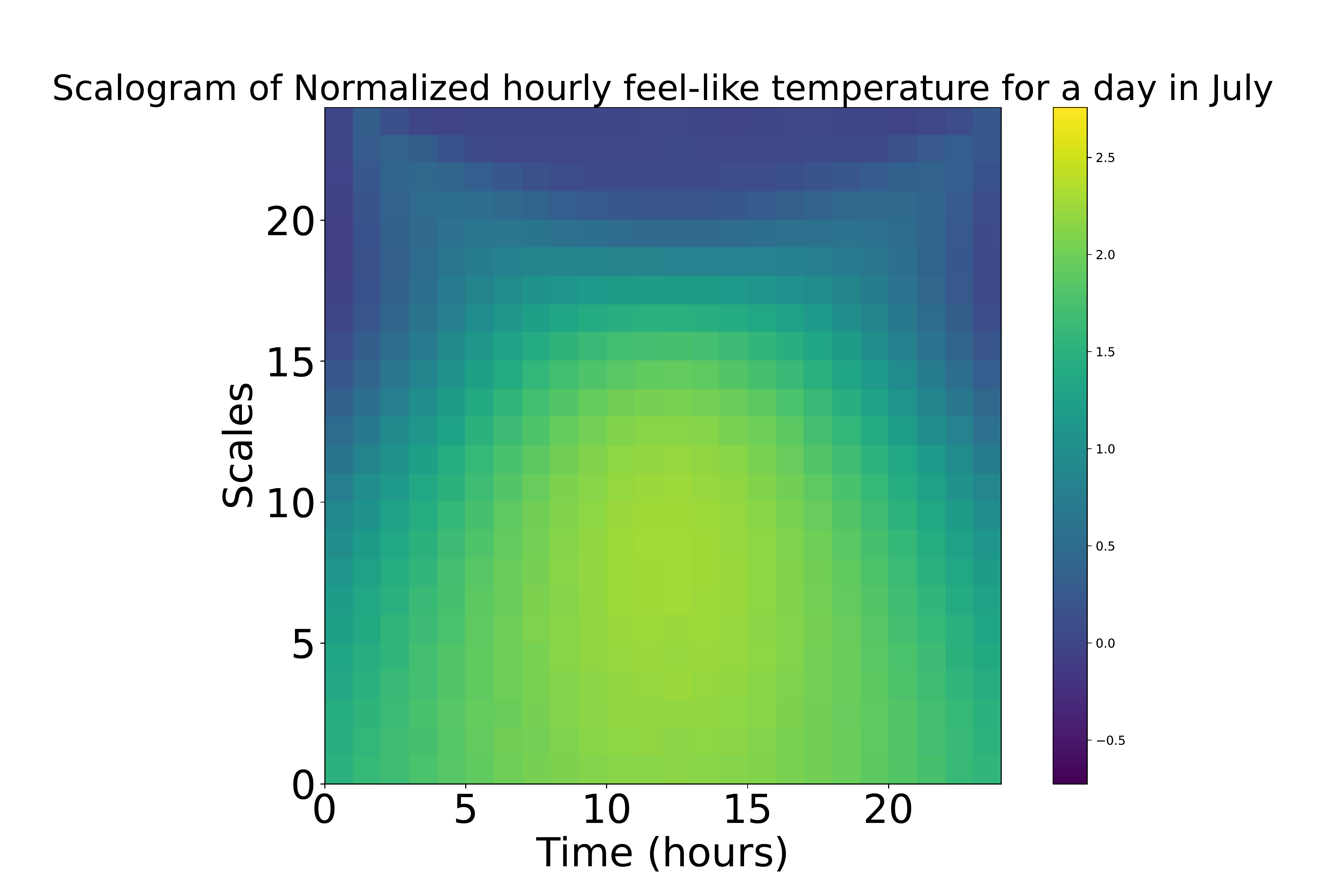} 
        \caption{Wavelet scalogram of feel-like temperature for a day in July.} 
        \label{Fig:weather_scalo_july}
    \end{subfigure}
    \begin{subfigure}[h]{0.45\linewidth} 
        \centering
        \includegraphics[width=\linewidth]{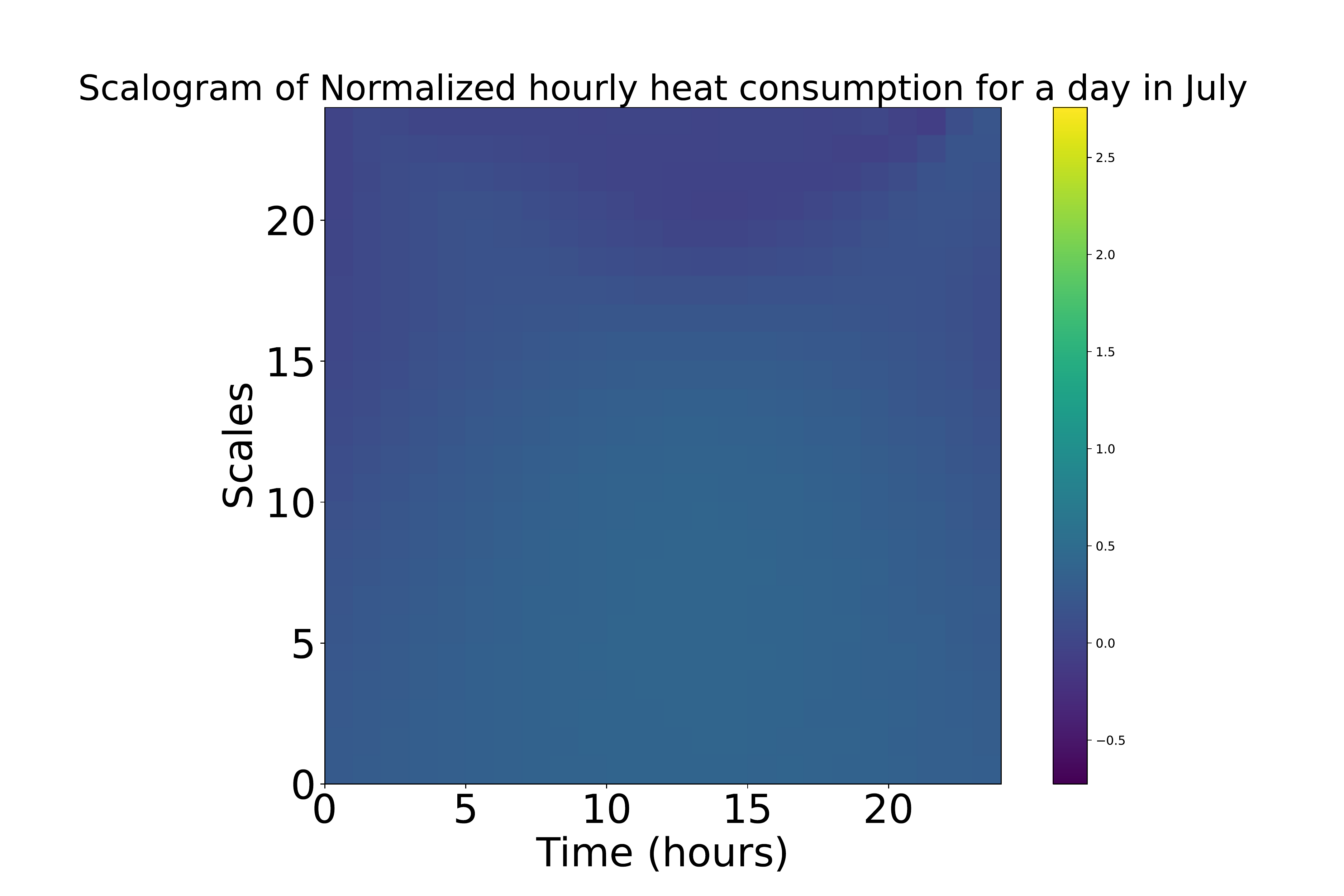} 
        \caption{Wavelet scalogram of heat consumption for a day in July}
        \label{Fig:Scalo_heat_july}
    \end{subfigure}
    \caption{Visualization of wavelet scalograms of the time series data used for model training.}
    \label{Fig:Scalograms}
\end{figure}

The wavelet scalograms shown in Figure \ref{Fig:Scalograms} depict the distribution of magnitude of wavelet transform coefficients over different scales and temporal resolution. Here the wavelet scalograms of weather information, i.e. feel-like temperature (Figure \ref{Fig:weather_scalo_jan}, Figure \ref{Fig:weather_scalo_july}) and those of heat consumption (Figure \ref{Fig:Scalo_heat_jan}, Figure \ref{Fig:Scalo_heat_july}) clearly shows the intuitive inverse relationship between weather and heat consumption. Clearly it is visible that not every time-scale component is having equal distribution of the magnitude of CWT coefficients, which the model intends to interpret, exploit and predict the future consumption vector.

\subsection{Impact of exogenous factors on heat demand}
\begin{figure}[htbp]
\centering
\includegraphics[width=\linewidth]{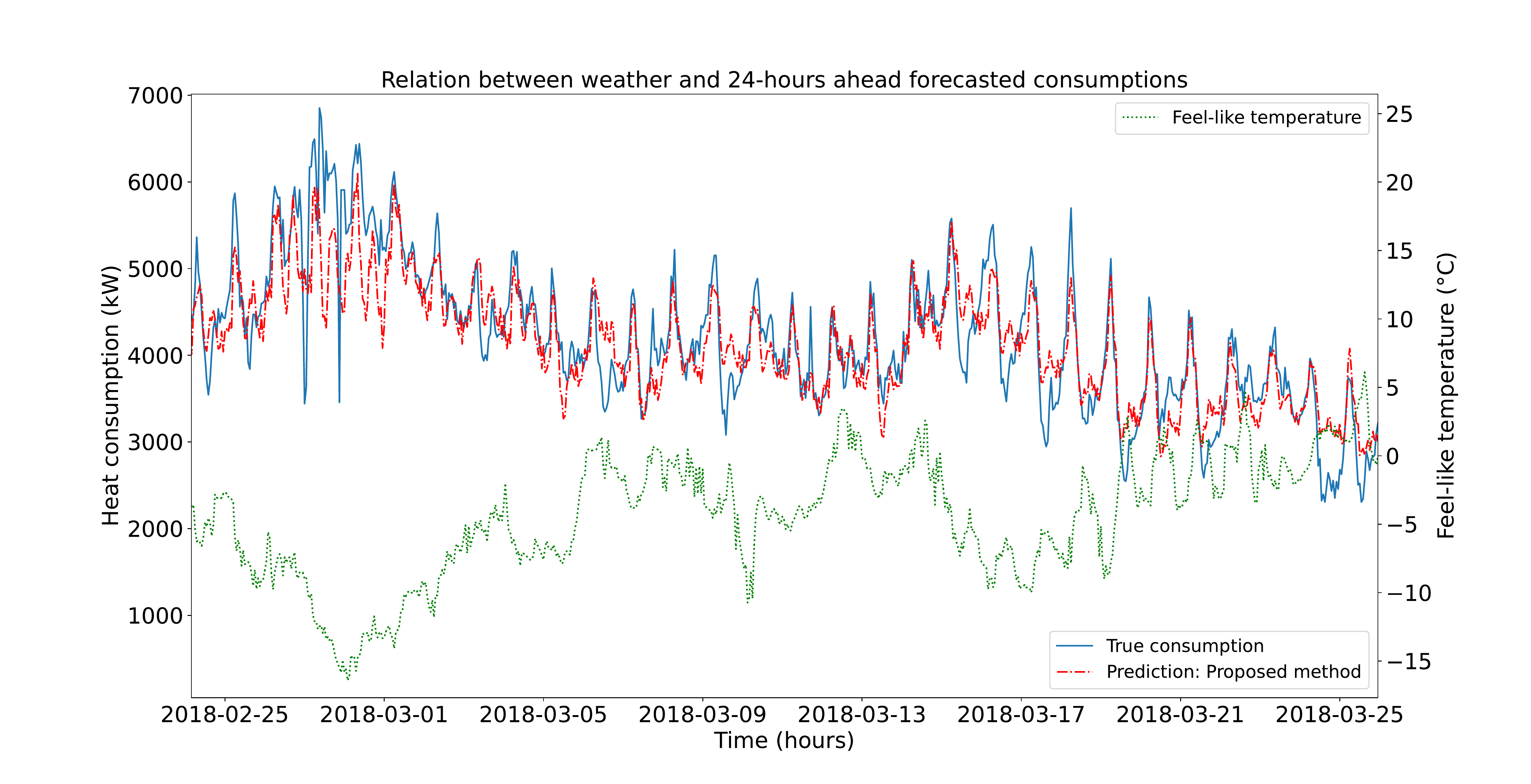}
\caption{Comparison between ground truth and predicted consumption and its relation with weather change over a period of one month}
\label{Fig:weatherheat}
\end{figure}

The Figure \ref{Fig:weatherheat} demonstrates the impact of weather on true and predicted heat consumption during early spring of the year which exhibits frequent weather fluctuation. But the heat consumption is not entirely steered by the weather but also with other factors, e.g. whether a day is a holiday or weekend. The behavior of the mass during weekends or holidays in a certain season of the year could still impact the heat consumption. 

\begin{figure}[htbp]
\centering
\includegraphics[width=\linewidth]{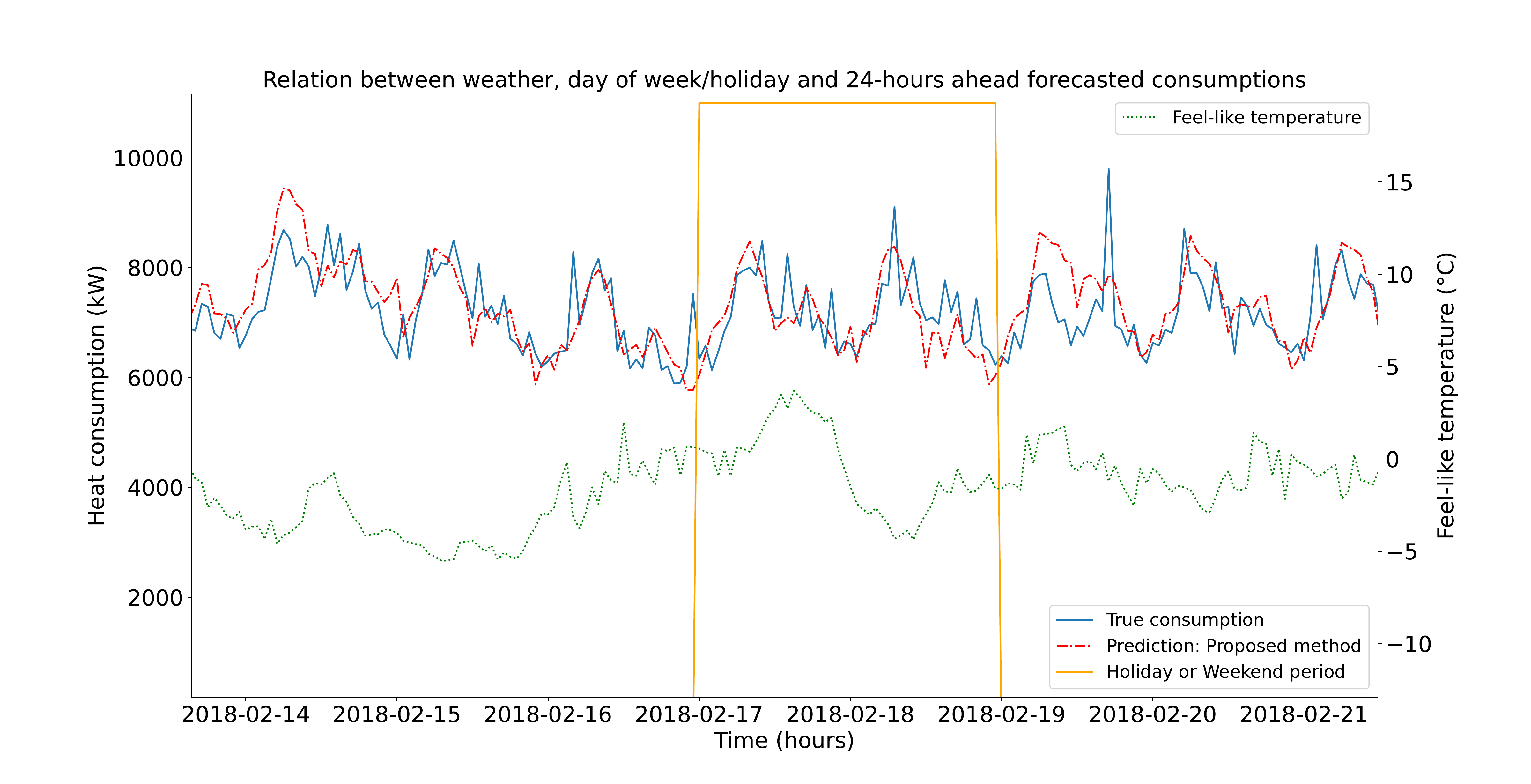}
\caption{Comparison between ground truth and predicted consumption and its relation with weather change and holiday over a period of one week}
\label{Fig:weatherholidayheat}
\end{figure}

As a proof of our claim that the model in our proposed method learns the complex non-linear relation among heat consumption, weather and day of week or public holiday, Figure \ref{Fig:weatherholidayheat} depicts the relationship among these factors and their impact on predicted heat consumption in zone 2, over the span of one week. The decreasing trend in feel-like temperature caused the decrease in true and predicted consumption from 14th to 16th February. Even though the weather constituted an increasing trend during the beginning of weekend, i.e. 17th February, the true and predicted heat consumption remained more or less same as it was on 15th February. It could be inferred that people in this region of Denmark during winter time tend to stay at home, yielding a rise to heat consumption which should have been lower as the feel-like temperature increased. 

\end{document}